# Gaussian Processes for Big Data

**James Hensman**[*]
Dept. Computer Science
The University of Sheffield
Sheffield, UK

**Nicolò Fusi**[*]
Dept. Computer Science
The University of Sheffield
Sheffield, UK

**Neil D. Lawrence**[*]
Dept. Computer Science
The University of Sheffield
Sheffield, UK

## Abstract

We introduce stochastic variational inference for Gaussian process models. This enables the application of Gaussian process (GP) models to data sets containing millions of data points. We show how GPs can be variationally decomposed to depend on a set of globally relevant inducing variables which factorize the model in the necessary manner to perform variational inference. Our approach is readily extended to models with non-Gaussian likelihoods and latent variable models based around Gaussian processes. We demonstrate the approach on a simple toy problem and two real world data sets.

## 1 Introduction

Gaussian processes [GPs, Rasmussen and Williams, 2006] are perhaps the dominant approach for inference on functions. They underpin a range of algorithms for regression, classification and unsupervised learning. Unfortunately, when applying a Gaussian process to a data set of size $n$ exact inference has complexity $\mathcal{O}(n^3)$ with storage demands of $\mathcal{O}(n^2)$. This hinders the application of these models for many domains. In particular, large spatiotemporal data sets, video, large social network data (e.g. from Facebook), population scale medical data sets, models that correlate across multiple outputs or tasks (for these models complexity is $\mathcal{O}(n^3p^3)$ and storage is $\mathcal{O}(n^2p^2)$ where $p$ is the number of outputs or tasks). Collectively we can think of these applications as belonging to the domain of 'big data'.

Traditionally in Gaussian process a large data set is one that contains over a few thousand data points.

Even to accommodate these data sets, various approximate techniques are required. One approach is to partition the data set into separate groups [e.g. Snelson and Ghahramani, 2007, Urtasun and Darrell, 2008]. An alternative is to build a low rank approximation to the covariance matrix based around 'inducing variables' [see e.g. Csató and Opper, 2002, Seeger et al., 2003, Quiñonero Candela and Rasmussen, 2005, Titsias, 2009]. These approaches lead to a computational complexity of $\mathcal{O}(nm^2)$ and storage demands of $\mathcal{O}(nm)$ where $m$ is a user selected parameter governing the number of inducing variables. However, even these reduced storage are prohibitive for big data, where $n$ can be many millions or billions. For parametric models, stochastic gradient descent is often applied to resolve this storage issue, but in the GP domain, it hasn't been clear how this should be performed. In this paper we show how recent advances in variational inference [Hensman et al., 2012, Hoffman et al., 2012] can be combined with the idea of inducing variables to develop a practical algorithm for fitting GPs using stochastic variational inference (SVI).

## 2 Sparse GPs Revisited

We start with a succinct rederivation of the variational approach to inducing variables of Titsias [2009]. This allows us to introduce notation and derive expressions which allow for the formulation of a SVI algorithm.

Consider a data vector[1] $\mathbf{y}$, where each entry $y_i$ is a noisy observation of the function $f(\mathbf{x}_i)$, for all the points $\mathbf{X} = \{\mathbf{x}_i\}_{i=1}^n$. We consider the noise to be independent Gaussian with precision $\beta$. Introducing a Gaussian process prior over $f(\cdot)$, let the vector $\mathbf{f}$ contain values of the function at the points $\mathbf{X}$. We shall also introduce a set of *inducing variables*: let the vector $\mathbf{u}$ contain values of the function $f$ at the points $\mathbf{Z} = \{\mathbf{z}_i\}_{i=1}^m$ which live in the same space as $\mathbf{X}$. Us-

---

[*]Also at Sheffield Institute for Translational Neuroscience, SITraN

[1]Our derivation trivially extends to multiple independent output dimensions, but we omit them here for clarity.

ing standard Gaussian process methodologies, we can write

$$p(\mathbf{y} \mid \mathbf{f}) = \mathcal{N}\left(\mathbf{y}|\mathbf{f}, \beta^{-1}\mathbf{I}\right),$$
$$p(\mathbf{f} \mid \mathbf{u}) = \mathcal{N}\left(\mathbf{f}|\mathbf{K}_{nm}\mathbf{K}_{mm}^{-1}\mathbf{u}, \widetilde{\mathbf{K}}\right),$$
$$p(\mathbf{u}) = \mathcal{N}\left(\mathbf{u}|\mathbf{0}, \mathbf{K}_{mm}\right),$$

where $\mathbf{K}_{mm}$ is the covariance function evaluated between all the inducing points and $\mathbf{K}_{nm}$ is the covariance function between all inducing points and training points and we have defined with $\widetilde{\mathbf{K}} = \mathbf{K}_{nn} - \mathbf{K}_{nm}\mathbf{K}_{mm}^{-1}\mathbf{K}_{mn}$.

We first apply Jensen's inequality on the conditional probability $p(\mathbf{y} \mid \mathbf{u})$:

$$\log p(\mathbf{y} \mid \mathbf{u}) = \log \langle p(\mathbf{y} \mid \mathbf{f}) \rangle_{p(\mathbf{f} \mid \mathbf{u})}$$
$$\geq \langle \log p(\mathbf{y} \mid \mathbf{f}) \rangle_{p(\mathbf{f} \mid \mathbf{u})} \triangleq \mathcal{L}_1. \quad (1)$$

where $\langle \cdot \rangle_{p(x)}$ denotes an expectation under $p(x)$. For Gaussian noise taking the expectation inside the log is tractable, but it results in an expression containing $\mathbf{K}_{nn}^{-1}$, which has a computational complexity of $\mathcal{O}(n^3)$. Bringing the expectation outside the log gives a lower bound, $\mathcal{L}_1$, which can be computed with has complexity $\mathcal{O}(m^3)$. Further, when $p(\mathbf{y}|\mathbf{f})$ factorises across the data,

$$p(\mathbf{y}|\mathbf{f}) = \prod_{i=1}^{n} p(y_i|f_i),$$

then this lower bound can be shown to be separable across $\mathbf{y}$ giving

$$\exp(\mathcal{L}_1) = \prod_{i=1}^{n} \mathcal{N}\left(y_i|\mu_i, \beta^{-1}\right) \exp\left(-\frac{1}{2}\beta \tilde{k}_{i,i}\right) \quad (2)$$

where $\boldsymbol{\mu} = \mathbf{K}_{nm}\mathbf{K}_{mm}^{-1}\mathbf{u}$ and $\tilde{k}_{i,i}$ is the $i$th diagonal element of $\widetilde{\mathbf{K}}$. Note that the difference between our bound and the original log likelihood is given by the Kullback Leibler (KL) divergence between the posterior over the mapping function given the data and the inducing variables and the posterior of the mapping function given the inducing variables only,

$$\mathrm{KL}\left(p(\mathbf{f}|\mathbf{u}) \,\|\, p(\mathbf{f}|\mathbf{u}, \mathbf{y})\right).$$

This KL divergence is minimized when there are $m = n$ inducing variables and they are placed at the training data locations. This means that $\mathbf{u} = \mathbf{f}$, $\mathbf{K}_{mm} = \mathbf{K}_{nm} = \mathbf{K}_{nn}$ meaning that $\widetilde{\mathbf{K}} = \mathbf{0}$. In this case we recover $\exp(\mathcal{L}_1) = p(\mathbf{y}|\mathbf{f})$ and the bound becomes equality because $p(\mathbf{f}|\mathbf{u})$ is degenerate. However, since $m = n$ and that there would be no computational or storage advantage from the representation. When $m < n$ the bound can be maximised with respect to $\mathbf{Z}$ (which are variational parameters). This minimises the KL divergence and ensures that $\mathbf{Z}$ are distributed amongst the training data $\mathbf{X}$ such that all $\tilde{k}_{i,i}$ are small. In practice this means that the expectations in (1) are only taken across a narrow domain ($\tilde{k}_{i,i}$ is the marginal variance of $p(f_i|\mathbf{u})$), keeping Jensen's bound tight.

Before deriving the expressions for stochastic variational inference using $\mathcal{L}_1$, we recover the bound of Titsias [2009] by marginalising the inducing variables,

$$\log p(\mathbf{y} \mid \mathbf{X}) = \log \int p(\mathbf{y} \mid \mathbf{u}) p(\mathbf{u}) \, \mathrm{d}\mathbf{u}$$
$$\geq \log \int \exp\{\mathcal{L}_1\} p(\mathbf{u}) \, \mathrm{d}\mathbf{u} \triangleq \mathcal{L}_2, \quad (3)$$

which with some linear algebraic manipulation leads to

$$\mathcal{L}_2 = \log \mathcal{N}\left(\mathbf{y}|\mathbf{0}, \mathbf{K}_{nm}\mathbf{K}_{mm}^{-1}\mathbf{K}_{mn} + \beta^{-1}\mathbf{I}\right) - \frac{1}{2}\beta \mathrm{tr}\left(\widetilde{\mathbf{K}}\right),$$

matching the result of Titsias, with the implicit approximating distribution $q(\mathbf{u})$ having precision

$$\boldsymbol{\Lambda} = \beta \mathbf{K}_{mm}^{-1} \mathbf{K}_{mn} \mathbf{K}_{nm} \mathbf{K}_{mm}^{-1} + \mathbf{K}_{mm}^{-1}$$

and mean

$$\hat{\mathbf{u}} = \beta \boldsymbol{\Lambda}^{-1} \mathbf{K}_{mm}^{-1} \mathbf{K}_{mn} \mathbf{y}.$$

## 3 SVI for GPs

One of the novelties of the Titsias bound was that, rather than explicitly representing a variational distribution for $q(\mathbf{u})$, these variables are 'collapsed' [Hensman et al., 2012]. However, for stochastic variational inference to work on Gaussian processes, it turns out we need to maintain an explicit representation of these inducing variables.

Stochastic variational inference (SVI) allows variational inference for very large data sets, but it can only be applied to probabilistic models which have a set of *global* variables, and which factorise in the observations and latent variables as Figure 1(a). Gaussian Processes do not have global variables and exhibit no such factorisation (Figure 1(b)). By introducing inducing variables $\mathbf{u}$, we have an appropriate model for SVI (Figure 1(c)). Unfortunately, marginalising $\mathbf{u}$ re-introduces dependencies between the observations, and eliminates the global parameters. In the following, we derive a lower bound on $\mathcal{L}_2$ which includes an explicit variational distribution $q(\mathbf{u})$, enabling SVI. We then derive the required natural gradients and discuss how latent variables might be used.

Because there are a fixed number of inducing variables (specified by the user at algorithm design time) we can perform stochastic variational inference, greatly increasing the size of data sets to which we can apply Gaussian processes.

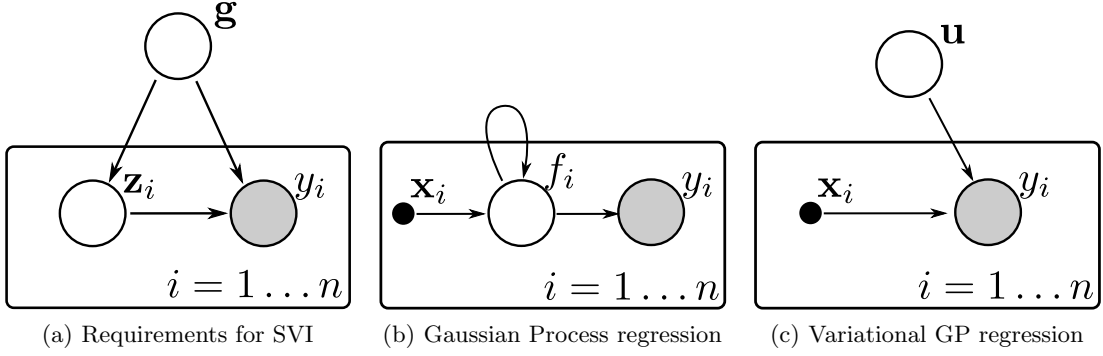

(a) Requirements for SVI  (b) Gaussian Process regression  (c) Variational GP regression

Figure 1: Graphical models showing (a) the reqired form for a probabilistic model for SVI (reproduced from [Hoffman et al., 2012]), with *global* variables $\mathbf{g}$ and latent variables $\mathbf{z}$. (b) The graphical model corresponding to Gaussian process regression, where connectivity between the values of the function $f_i$ is denoted by a loop around the plate. (c) The graphical model corresponding to the sparse GP model, with inducing variables $\mathbf{u}$ working as global variables, and the term $\mathcal{L}_1$ acting as $\log p(y_i \,|\, \mathbf{u}, \mathbf{x}_i)$. Marginalisation of $\mathbf{u}$ leads to the variational DTC formulation, introducing dependencies between the observations.

### 3.1 Global Variables

To apply stochastic variational inference to a Gaussian process model, we must have a set of global variables. The variables $\mathbf{u}$ will perform this role, and we introduce a variational distribution $q(\mathbf{u})$, and use it to lower bound the quantity $p(\mathbf{y} \,|\, \mathbf{X})$.

$$\log p(\mathbf{y}\,|\,\mathbf{X}) \geq \langle \mathcal{L}_1 + \log p(\mathbf{u}) - \log q(\mathbf{u}) \rangle_{q(\mathbf{u})} \triangleq \mathcal{L}_3.$$

From the above we know that the optimal distribution is Gaussian, and we parametrise it as $q(\mathbf{u}) = \mathcal{N}(\mathbf{u}\,|\,\mathbf{m}, \mathbf{S})$. The bound $\mathcal{L}_3$ becomes

$$\mathcal{L}_3 = \sum_{i=1}^n \left\{ \log \mathcal{N}\left(y_i | \mathbf{k}_i^\top \mathbf{K}_{mm}^{-1} \mathbf{m}, \beta^{-1}\right) \right.$$
$$\left. - \tfrac{1}{2}\beta \widetilde{k}_{i,i} - \tfrac{1}{2}\mathrm{tr}\left(\mathbf{S}\mathbf{\Lambda}_i\right) \right\}$$
$$- \mathrm{KL}\left(q(\mathbf{u}) \,\|\, p(\mathbf{u})\right) \qquad (4)$$

with $\mathbf{k}_i$ being a vector of the $i^{\text{th}}$ column of $\mathbf{K}_{mn}$ and $\mathbf{\Lambda}_i = \beta \mathbf{K}_{mm}^{-1} \mathbf{k}_i \mathbf{k}_i^\top \mathbf{K}_{mm}^{-1}$. The gradients of $\mathcal{L}_3$ with respect to the parameters of $q(\mathbf{u})$ are

$$\frac{\partial \mathcal{L}_3}{\partial \mathbf{m}} = \beta \mathbf{K}_{mm}^{-1} \mathbf{K}_{mn} \mathbf{y} - \mathbf{\Lambda}\mathbf{m},$$
$$\frac{\partial \mathcal{L}_3}{\partial \mathbf{S}} = \tfrac{1}{2}\mathbf{S}^{-1} - \tfrac{1}{2}\mathbf{\Lambda}. \qquad (5)$$

Setting the derivatives to zero recovers the optimal solution found in the previous section, namely $\mathbf{S}=\mathbf{\Lambda}^{-1}$, $\mathbf{m}=\hat{\mathbf{u}}$. It follows that $\mathcal{L}_2 \geq \mathcal{L}_3$, with equality at this unique maximum.

The key propery of $\mathcal{L}_3$ is that is can be written as a sum of $n$ terms, each corresponding to one input-output pair $\{\mathbf{x}_i, y_i\}$: we have induced the necessary factorisation to perform stochastic gradient methods on the distribution $q(\mathbf{u})$.

### 3.2 Natural Gradients

Stochastic variational inference works by taking steps in the direction of the approximate *natural* gradient $\widetilde{\mathbf{g}}(\boldsymbol{\theta})$, which is given by the usual gradient re-scaled by the inverse Fisher information: $\widetilde{\mathbf{g}}(\boldsymbol{\theta}) = G(\boldsymbol{\theta})^{-1} \frac{\partial \mathcal{L}}{\partial \boldsymbol{\theta}}$. To work with the natural gradients of the distribution $q(\mathbf{u})$, we first recall the canonical and expectation parameters

$$\boldsymbol{\theta}_1 = \mathbf{S}^{-1}\mathbf{m}, \quad \boldsymbol{\theta}_2 = -\tfrac{1}{2}\mathbf{S}^{-1}$$

and

$$\boldsymbol{\eta}_1 = \mathbf{m}, \quad \boldsymbol{\eta}_2 = \mathbf{m}\mathbf{m}^\top + \mathbf{S}.$$

In the exponential family, properties of the Fisher information reveal the following simplification of the natural gradient [Hensman et al., 2012],

$$\widetilde{\mathbf{g}}(\boldsymbol{\theta}) = G(\boldsymbol{\theta})^{-1} \frac{\partial \mathcal{L}_3}{\partial \boldsymbol{\theta}} = \frac{\partial \mathcal{L}_3}{\partial \boldsymbol{\eta}}. \qquad (6)$$

A step of length $\ell$ in the natural gradient direction, using $\boldsymbol{\theta}_{(t+1)} = \boldsymbol{\theta}_{(t)} + \ell \frac{\mathrm{d}\mathcal{L}_3}{\mathrm{d}\boldsymbol{\eta}}$, yields

$$\boldsymbol{\theta}_{2(t+1)} = -\tfrac{1}{2}\mathbf{S}_{(t+1)}^{-1}$$
$$= -\tfrac{1}{2}\mathbf{S}_{(t)}^{-1} + \ell\left(-\tfrac{1}{2}\mathbf{\Lambda} + \tfrac{1}{2}\mathbf{S}_{(t)}^{-1}\right),$$
$$\boldsymbol{\theta}_{1(t+1)} = \mathbf{S}_{(t+1)}^{-1}\mathbf{m}_{(t+1)}$$
$$= \mathbf{S}_{(t)}^{-1}\mathbf{m}_{(t)} + \ell\left(\beta \mathbf{K}_{mm}^{-1}\mathbf{K}_{mn}\mathbf{y} - \mathbf{S}_{(t)}^{-1}\mathbf{m}_{(t)}\right),$$

and taking a step of unit length then recovers the same solution as above by either (3) or (5). This confirms the result discussed in Hensman et al. [2012], Hoffman et al. [2012], that taking this unit step is the same as performing a VB update. We can now obtain stochastic approximations to the natural gradient by considering the data either individually or in mini-batches.

We note the convenient result that the natural gradient for $\boldsymbol{\theta}_2$ is positive definite (note $\boldsymbol{\Lambda} = \mathbf{K}_{mm}^{-1} + \sum_i \boldsymbol{\Lambda}_i$). This means that taking a step in that direction always leads to a positive definite matrix, and our implementation need not parameterise $\mathbf{S}$ in any way so as to ensure positive-definiteness, *cf.* standard gradient approaches on covariance matrices.

To optimise the kernel hyper-parameters and noise precision $\beta$, we take derivatives of the bound $\mathcal{L}_3$ and perform standard stochastic gradient descent alongside the variational parameters. An illustration is presented in Figure 2.

### 3.3 Latent Variables

The above derivations enable online learning for Gaussian process *regression* using SVI. Several GP based models involve inference of $\mathbf{X}$, such as the GP latent variable model [Lawrence, 2005, Titsias and Lawrence, 2010] and its extensions [e.g. Damianou et al., 2011, 2012].

To perform stochastic variational inference with latent variables, we require a factorisation as illustrated by Figure 1(a): this factorisation is provided by (4). To get a model like the Bayesian GPLVM, we need a lower bound on $\log p(\mathbf{y})$. In Titsias and Lawrence [2010] this was achieved through approximate marginalisation of $\mathcal{L}_2$, w.r.t. $\mathbf{X}$, which leads to an expression depending only on the parameters of $q(\mathbf{X})$. However this formulation scales poorly, and the variables of the optimisation are closely connected due to the marginalisation of $\mathbf{u}$. The above enables a lower bound to which SVI is immediately applicable:

$$\log p(\mathbf{y}) = \log \int p(\mathbf{y} \mid \mathbf{X}) p(\mathbf{X}) \, \mathrm{d}\mathbf{X}$$
$$\geq \int q(\mathbf{X}) \{\mathcal{L}_3 + \log p(\mathbf{X}) - \log q(\mathbf{X})\} \, \mathrm{d}\mathbf{X}.$$

It is straightforward to introduce $p$ output dimensions for the data $\mathbf{Y}$, and following Titsias and Lawrence [2010], we use a factorising normal distribution $q(\mathbf{X}) = \prod_{i=1}^n q(\mathbf{x}_i)$. The relevant expectations of $\mathcal{L}_3$ are tractable for various choices of covariance function.

To perform SVI in this model, we now alternate between selecting a minibatch of data, and optimising the relevant variables of $q(\mathbf{X})$ with $q(\mathbf{u})$ fixed, and updating $q(\mathbf{u})$ using the approximate natural gradient. We note that the form of (4) means that each of the latent variable distributions may be updated individually, enabling parallelisation across the minibatch.

### 3.4 Non-Gaussian likelihoods

Another advantage of the factorisation of (4) is that it enables a simple routine for inference with non-Gaussian likelihoods. The usual procedure for fitting GPs with non-Gaussian likelihoods is to approximate the likelihood using either a local variational lower bound [Gibbs and MacKay, 2000], or by expectation propagation [Kuss and Rasmussen, 2005]. These approximations to the likelihood are required because of the connections between the variables $\mathbf{f}$.

In $\mathcal{L}_3$, the bound factorises in such a way that some non-Gaussian likelihoods may be marginalised *exactly*, given the existing approximations. To see this, consider that we are presented not with the vector $\mathbf{y}$, but by a binary vector $\mathbf{t}$ with $t_i \in \{0, 1\}$, and the likelihood $p(\mathbf{t} \mid \mathbf{y}) = \prod_{i=1}^n \sigma(y_i)^{t_i}(1-\sigma(y_i))^{(1-t_i)}$, as in the case of classification. We can bound the marginal likelihood using $p(\mathbf{t} \mid \mathbf{X}) \geq \int p(\mathbf{t} \mid \mathbf{y}) \exp\{\mathcal{L}_3\} \, \mathrm{d}\mathbf{y}$ which involves $n$ independent one dimensional integrals due to the factorising nature of $\mathcal{L}_3$. For the probit likelihood each of these integrals is tractable.

This kind of approximation, where the likelihood is integrated exactly is amenable to SVI in the same manner as the regression case above through computation of the natural gradient.

## 4 Experiments

### 4.1 Toy Data

To demonstrate our algorithm we begin with two simple toy datasets based on sinusoidal functions. In the first experiment we show how the approximation converges towards the true solution as mini-batches are included. Figure 2 shows the nature of our approximation: the variational approximation to the inducing function variables is shown.

The second toy problem (Figure 3) illustrates the convergence of the algorithm on a two dimensional problem, again based on sinusoids. Here, we start with a random initialisation for $q(\mathbf{u})$, and the model converges after 2000 iterations. We found empirically that holding the covariance parameters fixed for the first epoch results in more reliable convergence, as can be seen in Figure 4

### 4.2 UK Apartment Price Data

Our first large scale Gaussian process models the changing cost of apartments in the UK. We downloaded the monthly price paid data for the period February to October 2012 from `http://data.gov.uk/dataset/`

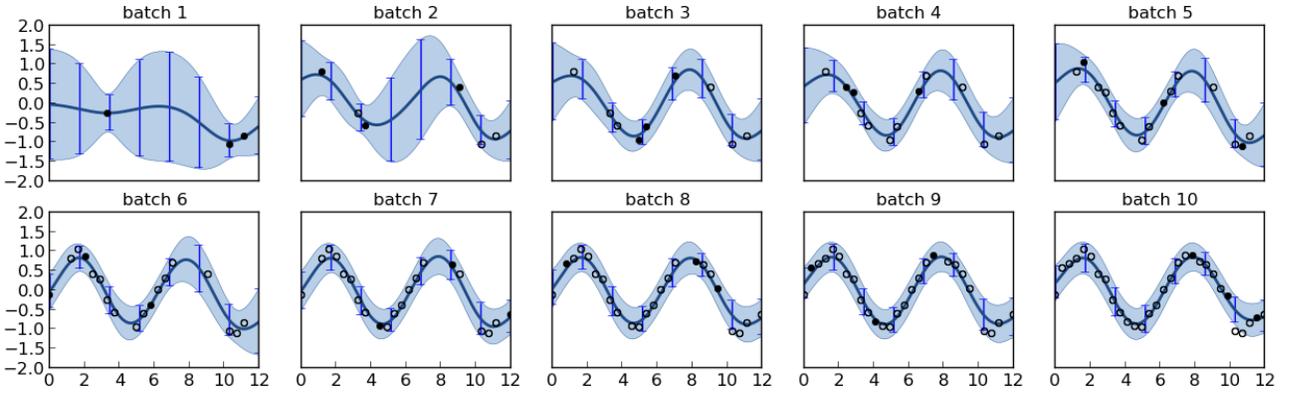

Figure 2: Stochastic variational inference on a trivial GP regression problem. Each pane shows the posterior of the GP after a batch of data, marked as solid points. Previoulsy seen (and discarded) data are marked as empty points, the distribution $q(\mathbf{u})$ is represented by vertical errorbars.

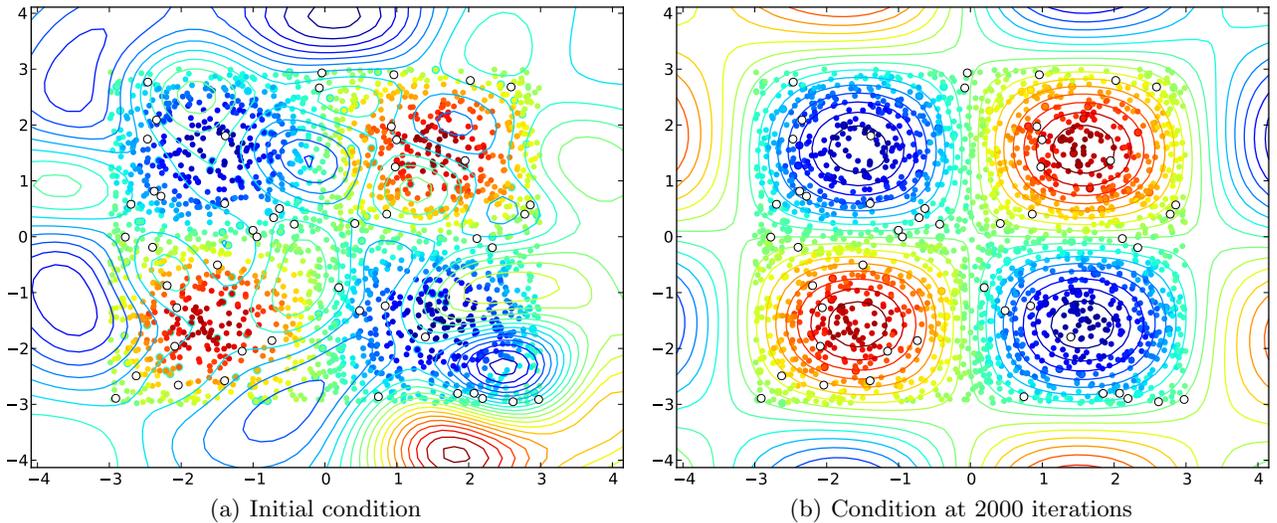

(a) Initial condition

(b) Condition at 2000 iterations

Figure 3: A two dimensional toy demo, showing the initial condition and final condition of the model. Data are marked as colored points, and the model's prediction is shown as (similarly colored) contour lines. The positions of the inducing variables are marked as empty circles.

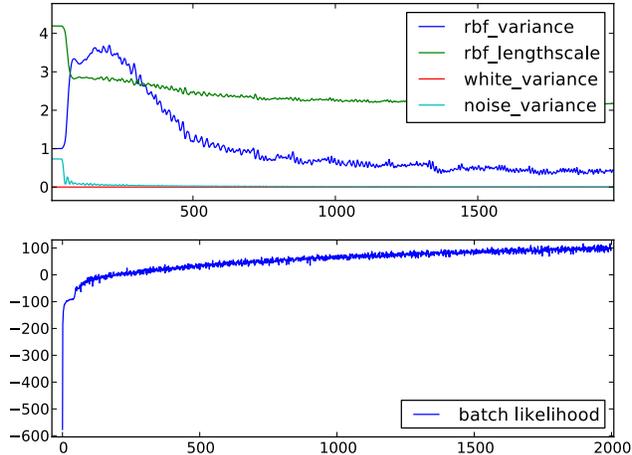

Figure 4: Convergence of the SVIGP algorithm on the two dimensional toy data

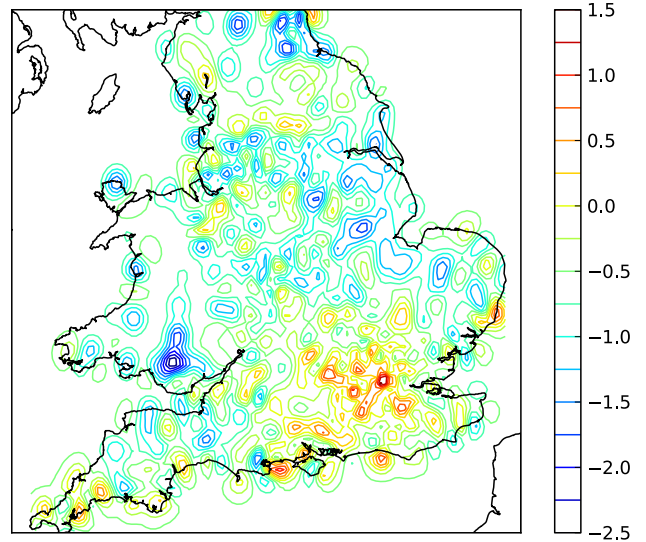

Figure 5: Variability of apartment price (logarithmically!) throughout England and Wales.

`land-registry-monthly-price-paid-data/`, which covers England and Wales, and filtered for apartments. This resulted in a data set with 75,000 entries, which we cross referenced against a postcode database to get lattitude and longitude, on which we regressed the normalised logarithm of the apartment prices.

Randomly selecting 10,000 data as a test set, we build a GP as described with a covariance function $k(\cdot, \cdot)$ consisting of four parts: two squared exponential covariances, initialised with different length scales were used to account for national and regional variations in property prices, a constant (or 'bias') term allowed for non-zero mean data, and a noise variance accounted for variation that could not be modelled using simply latitude and longitude.

We selected 800 inducing input sites using a $k$-means algorithm, and optimised the parameters of the covariance function alongside the variational parameters. We performed some manual tuning of the learning rates: empirically we found that the step length should be much higher for the variational parameters of $q(\mathbf{u})$ than for the values of the covariance function parameters. We used 0.01 and $1 \times 10^{-5}$. Also, we included a momentum term for the covariance function parameters (set to 0.9). We tried including momentum terms for the variational parameters, but we found this hindered performance. A large mini-batch size (1000) reduced the stochasticity of the gradient computations. We judged that the algorithm had converged after 750 iterations, as the stochastic estimate of the marginal lower bound on the marginal likelihood failed to increase further.

For comparison to our model, we constructed a series of GPs on subsets of the training data. Splitting the data into sets of 500, 800, 1000 and 1200, we fitted a GP with the same covariance function as our stochastic GP. Parameters of the covariance function were optimised using type-II maximum likelihood for each batch. Table 1 reports the mean squared error in our model's prediction of the held out prices, as well as the same for the random sub-set approach (along with two standard deviations of the inter-sub-set variability).

Table 1: Mean squared errors in predicting the log-apartment prices across England and Wales by lattitude and longitude

|  | Mean square Error |
|---|---|
| SVIGP | **0.426** |
| Random sub-set (N=500) | 0.522 +/- 0.018 |
| Random sub-set (N=800) | 0.510 +/- 0.015 |
| Random sub-set (N=1000) | 0.503 +/- 0.011 |
| Random sub-set (N=1200) | 0.502 +/- 1.012 |

### 4.3 Airline Delays

The second large scale dataset we considered consists of flight arrival and departure times for every commercial flight in the USA from January 2008 to April 2008. This dataset contains extensive information about almost 2 million flights, including the delay (in minutes) in reaching the destination. The average delay of a flight in the first 4 months of 2008 was of 30 minutes. Of course, much better estimates can be given by exploiting the enormous wealth of data available, but rich models are often overlooked in these cases due to the sheer size of the dataset. We randomly selected 800,000 datapoints [2], using a random subset of 700,000 samples to train the model and 100,000 to test it. We chose to include into our model 8 of the many variables available for this dataset: the age of the aircraft (number of years since deployment), distance that needs to be covered, airtime, departure time, arrival time, day of the week, day of the month and month.

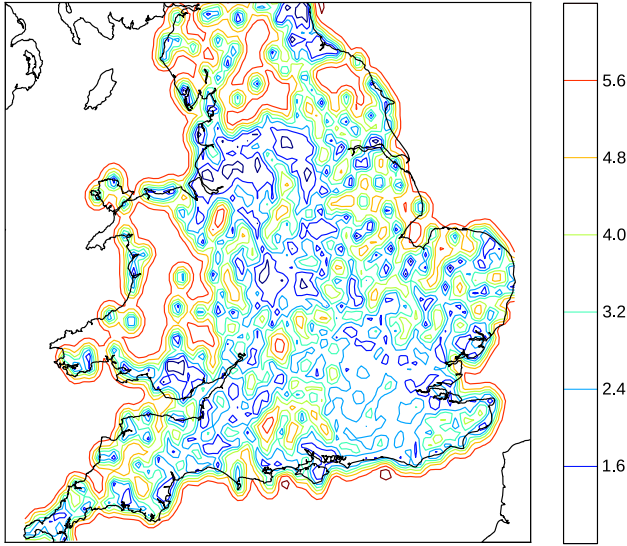

Figure 6: Posterior variance of apartment prices.

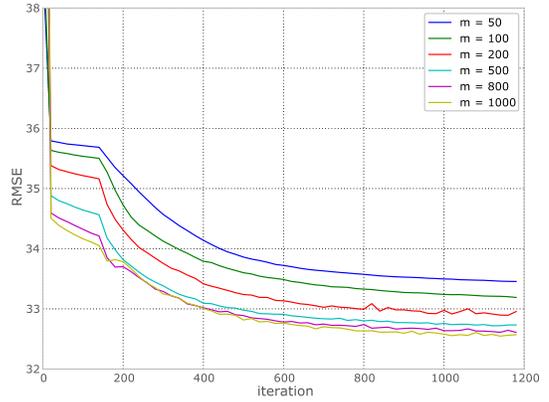

Figure 8: Root mean square errors for models with different numbers of inducing variables.

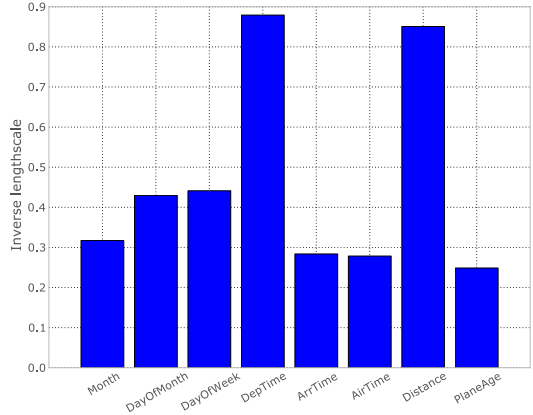

Figure 9: Automatic relevance determination parameters for the features used for predicting flight delays.

We built a Gaussian process with a squared exponential covariance function with a bias and noise term. In order to discard irrelevant input dimensions, we allowed a separate lengthscale for each input. For our experiments, we used $m = 1000$ inducing inputs and a mini-batch size of 5000. The learning rate for the variational parameters of $q(\mathbf{u})$ was set to 0.01, while the learning rate for the covariance function parameters was set to $1 \times 10^{-5}$. We also used a momentum term of 0.9 for the covariance parameters.

For the purpose of comparison, we fitted several GPs with an identical covariance function on subsets of the data. We split the data into sets of 800, 1000 and 1200 samples and optimised the parameters using type-II maximum likelihood. We repeated this procedure 10 times.

The left pane of Figure 7 shows the root mean squared error (RMSE) obtained by fitting GPs on subsets of the data. The right pane of figure 7 shows the RMSE obtained by fitting 10 SVI GPs as a function of the iteration. The individual runs are shown in light gray, while the blue line shows the average RMSE across runs.

One of the main advantages of the approach presented here is that the computational complexity is independent from the number of samples $n$. This allowed us to use a much larger number of inducing inputs than has traditionally been possible. Conventional sparse GPs have a computational complexity of $\mathcal{O}(nm^2)$, so for large $n$ the typical upper bound for $m$ is between 50 and 100. The impact on the prediction performance is quite significant, as highlighted in Figure 8, where we fit several SVI GPs using different numbers of inducing inputs.

Looking at the inverse lengthscales in Figure 9, it's possible to get a better idea of the relevance of the different features available in this dataset. The most relevant variable turned out to be the time of departure of the flight, closely followed by the distance that needs to be covered. Distance and airtime should in theory be correlated, but they have very different relevances. This can be intuitively explained by considering that on longer flights it's easier to make up for delays at departure.

---

[2] Subsampling wasn't technically necessary, but we didn't want to overburden the memory of a shared compute node just before a submission deadline.

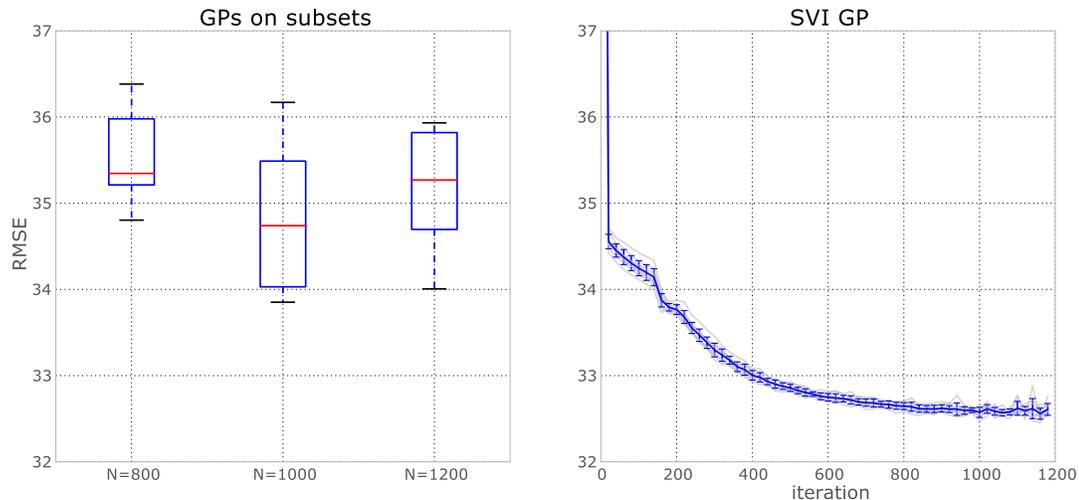

Figure 7: Root mean squared errors in predicting flight delays using information about the flight.

## 5 Discussion

We have presented a method for inference in Gaussian process models using stochastic variational inference. These expressions allow for the transfer of a multitude of Gaussian process techniques to big data.

We note several interesting results. First, the our derivation disusses the bound on $p(\mathbf{y}\,|\,\mathbf{u})$ in detail, showing that it becomes tight when $\mathbf{Z} = \mathbf{X}$.

Also, we have that there is a unique solution for the parameters of $q(\mathbf{u})$ such that the bound associated with the standard variational sparse GP [Titsias, 2009] is recovered.

Further, since the complexity of our model is now $\mathcal{O}(m^3)$ rather than $\mathcal{O}(nm^2)$, we are free to increase $m$ to much greater values than the sparse GP representation. The effect of this is that we can have much richer models: for a squared exponential covariance function, we have far more basis-functions with which to model the data. In our UK apartment price example, we had no difficulty setting $m$ to 800, much higher than experience tells us is feasible with the sparse GP.

The ability to increase the number of inducing variables and the applicability to unlimited data make our method suitable for multiple output GPs [Álvarez and Lawrence, 2011]. We have also briefly discussed how this framework fits with other Gaussian process based models such as the GPLVM and GP classification. We leave the details of these implementations to future work.

In all our experiments our algorithm was run on a single CPU using the GPy Gaussian process toolkit https://github.com/SheffieldML/GPy.